\documentclass[runningheads]{llncs}
\usepackage{comment}
\usepackage{hyperref}
\usepackage{graphicx}
\usepackage{multirow}
\begin{document}
\title{Chinese Word Sense Embedding with SememeWSD and Synonym Set}

\def\CICAISubNumber{135}  
\titlerunning{CICAI2022 submission ID \CICAISubNumber} 
\authorrunning{CICAI2022 submission ID \CICAISubNumber} 
\author{Yangxi Zhou \and Junping Du\thanks{Corresponding author: Junping Du (junpingdu@126.com)} \and Zhe Xue \and Ang Li \and Zeli Guan}
\institute{
Beijing Key Laboratory of Intelligent Telecommunication Software and Multimedia, School of Computer Science (National Pilot School of Software Engineering), \\
Beijing University of Posts and Telecommunications, \\
Beijing 100876, China}

\maketitle              

\begin{abstract}
Word embedding is a fundamental natural language processing task which can learn feature of words. However, most word embedding methods assign only one vector to a word, even if polysemous words have multi-senses. To address this limitation, we propose SememeWSD Synonym (SWSDS) model to assign a different vector to every sense of polysemous words with the help of word sense disambiguation (WSD) and synonym set in OpenHowNet. We use the SememeWSD model, an unsupervised word sense disambiguation model based on OpenHowNet, to do word sense disambiguation and annotate the polysemous word with sense id. Then, we obtain top 10 synonyms of the word sense from OpenHowNet and calculate the average vector of synonyms as the vector of the word sense. In experiments, We evaluate the SWSDS model on semantic similarity calculation with Gensim's wmdistance method. It achieves improvement of accuracy. We also examine the SememeWSD model on different BERT models to find the more effective model.

\keywords{Word Sense Disambiguation  \and Word Sense Embedding \and Synonym Set.}
\end{abstract}
\section{Introduction}
The purpose of word embedding is to learn the vector representation of a word in its context~\cite{ref_article1}. The learned word vectors can be used in subsequent natural language processing tasks, such as semantic similarity calculating ~\cite{ref_article2,ref_article3,ref_article4} and sentiment analysis~\cite{ref_article5,ref_article6,ref_article7}. The quality of word vectors learned by word embedding method can directly influence the performance of these tasks. Therefore, improving the performance of word embedding is a critical issue for natural language understanding.

\begin{figure}
\centering
\includegraphics[scale=0.8]{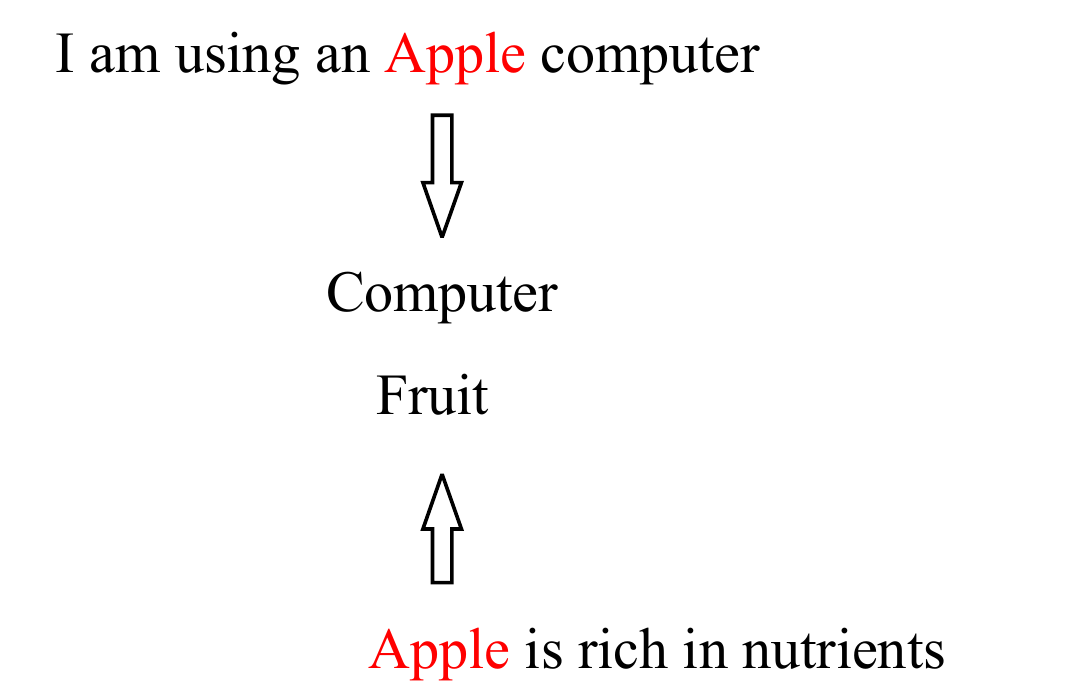}
\caption{Different senses of "Apple" in different contexts.} \label{apple}
\end{figure}

Researches show that polysemy accounts for about 42 percent of natural language. However, most traditional word embedding methods assume that each word corresponds to only one word vector, even if it is a polysemous word. It means that they are single-prototype word vectors. For example, word2vec (Mikolov et al., 2013)~\cite{ref_article8} is the most popular word embedding method that strikes a good balance between effectiveness and efficiency and assigns only one vector to each word. “Apple” is a typical polysemous word that has many senses. As suggested in Figure~\ref{apple}, on the one hand, “Apple” means “Computer” in the context “I am using an apple computer”. On the other hand, it means “Fruit” in the context “Apple is rich in nutrients”. As a result, single prototype word vectors cannot convey the different senses of polysemous words in different contexts.

\noindent To resolve issues mentioned above, Huang et al. (2012) ~\cite{ref_article9} present a multi-prototype model based on neural network. Furthermore, Chen et al. (2014)~\cite{ref_article10} combine word sense representation with word sense disambiguation and they make use of the synonyms set in WordNet~\cite{ref_article11}. Word sense disambiguation aims to determine the specific sense of polysemous words in the context. In reality, OpenHowNet is the most famous Chinese/English sememe knowledge base which also annotate each word sense with the synonym set~\cite{ref_article12}. From previous studies, we learn that word sense disambiguation is beneficial to word sense embedding (WSE) and synonym set can convey the sense of the polysemous word to a certain extent.

In this paper, we intend to further incorporate word sense disambiguation and synonym set in OpenHowNet into word sense embedding for Chinese. To prove its feasibility, we present a SememeWSD-Synonym (SWSDS) model, which disambiguates polysemous words and represents it by synonym set. In experiments, we further study the SememeWSD model on WSD dataset and we also evaluate our model on semantic similarity. The experiment results show that our model outperforms the baseline method. This means that our model can improve word sense embedding for subsequent tasks with the help of word sense disambiguation and synonym set.

The key contributions of this paper can be concluded as follows:

(1)	We built a polysemous words' dictionary based on OpenHowNet to identify polysemous words in a given sentence.

(2)	We conducted extensive experiments on SememeWSD and found a more effective BERT model for word sense disambiguation dataset.

(3)	To the best of our knowledge, this is the first work to utilize synonym set in OpenHowNet to improve word sense embedding.

\section{Related Work}
\subsection{Word Embedding}
With the rapid development of natural language processing, word embedding has become the most important method to obtain the effective representations of text words ~\cite{ref_article13,ref_article14,ref_article15,ref_article16}.

The one-hot representation is a method that represents each text word with a binary vector. As the construction is extremely simple, the one-hot representation has been widely used in some NLP tasks~\cite{ref_article17}. However, this representation method is likely to suffer from dimension curse ~\cite{ref_article18}.

To address this issue, Bengio et al.~\cite{ref_article18} presented the concept of word embedding. Afterwards, Mikolov et al.~\cite{ref_article8} proposed word2vec model based on neural network, including Skip-Gram and CBOW. It strikes a good balance between effectiveness and efficiency and assigns only one vector to each word. But the sense of a text word changes with the context. To solve the problem, Pennington et al. ~\cite{ref_article19} present GloVe to combine the merits of the matrix factorization and the prediction-based methods. Embeddings from Language Model (EMLO) is the implementation of the deep contextualized word representations proposed by Matthew et al.~\cite{ref_article20}. Huang et al.~\cite{ref_article9} utilize multi-prototype vector models to learn word embedding and arrange a distinct vector for each word sense. 

\subsection{Word Sense Disambiguation and Word Sense Embedding}
With the popularity of the natural language processing, word sense disambiguation has gradually emerged into our vision and becomes increasingly important to many tasks ~\cite{ref_article21,ref_article22,ref_article23,ref_article24,ref_article25}. It aims to identify word sense in a context which contains two approaches. Supervised methods utilize classifiers for word sense disambiguation~\cite{ref_article26}. It requires large-scale human annotation data.

Instead, knowledge-based methods base on large external knowledge bases to find the most likely sense of a word. Agirre et al.~\cite{ref_article27} present a WSD algorithm based on random walks over large Lexical Knowledge Bases. Ustalov et al. ~\cite{ref_article28} propose a model that first obtains sense embeddings from the word embeddings of the corresponding senses' synonyms in WordNet and then selects the sense that has the closest embedding similarity with the context. Chen et al. ~\cite{ref_article10} take synsets in WordNet into consideration and do word sense disambiguation and word sense representation learning concurrently. Hou et al.~\cite{ref_article29} propose an unsupervised HowNet-based word sense disambiguation model with the help of BERT model which is named SememeWSD. They also build an OpenHowNet-based WSD datasets.

In a word, many researchers have made progress in word embedding and word sense disambiguation recently~\cite{ref_article30,ref_article31,ref_article32,ref_article33,ref_article34,ref_article35,ref_article36}. In this paper, we utilize the SememeWSD to do word sense disambiguation and do word sense embedding with synonym set in OpenHowNet.
\section{Methodology}
We now describe the SWSDS model, our proposed approach for word sense embedding. We first give a brief introduction to synonyms, sememes and senses in OpenHowNet and SememeWSD. Then we present the architecture of our model and explain how the model works.
\subsection{OpenHowNet and SememeWSD}
OpenHowNet originates from HowNet, which is the most famous sememe knowledge base. It defines about 2,000 sememes and uses them to annotate senses of more than 100,000 Chinese words and phrases~\cite{ref_article12}. The senses of all the words in HowNet can be represented by a set of sememes. Smeme is defined as the minimum semantic unit of human language.

\begin{figure}
\centering
\includegraphics[scale=0.8]{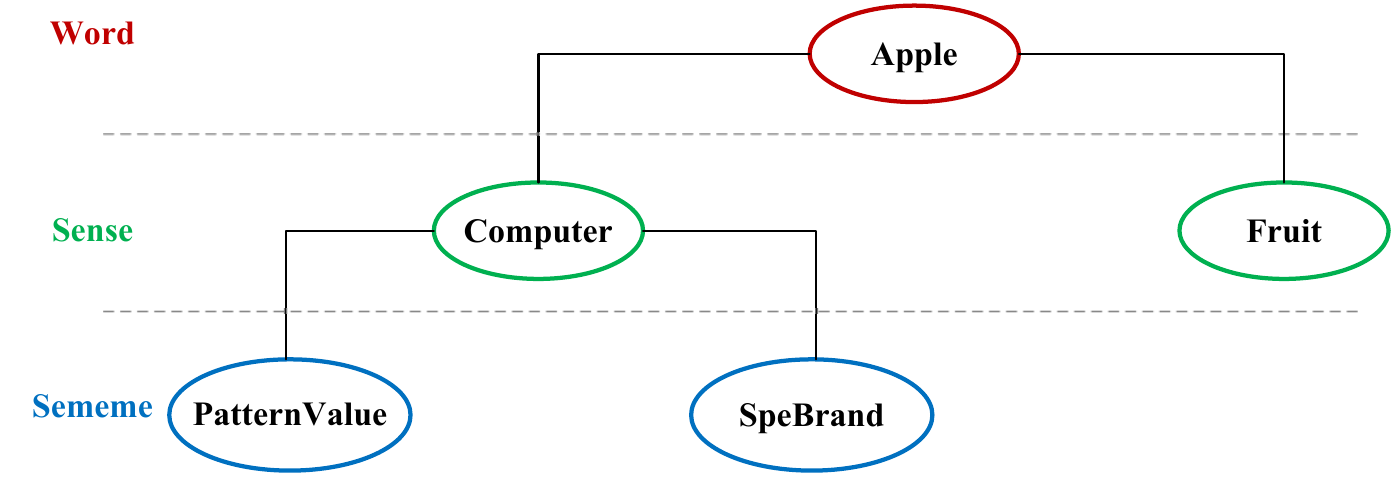}
\caption{Sememe annotations of the word “Apple” in OpenHowNet.} \label{sememe}
\end{figure}

Sememe annotations in OpenHowNet for each sense looks like a tree, as illustrated in Figure~\ref{sememe}. The word "Apple" has two senses including "Computer" and "Fruit". For the first sense "Computer", it is annotated with 3 sememes while the second is annotated with only 1 sememe.

\noindent According to the definition of sememe and the principle of OpenHowNet, sememes of a sense can convey its sense, and two senses with the same sememes are supposed to have the same meaning~\cite{ref_article29}. Therefore, in SememeWSD, the model first selects a set of substitution words which have a sense that has the same sememe annotations as the target sense and uses their average MLM prediction score to measure the possibility of the target sense in the context. Then SememeWSD chooses the target sense whose average MLM prediction score is the highest as the correct sense in the context.

\begin{figure}
\centering
\includegraphics[scale=0.8]{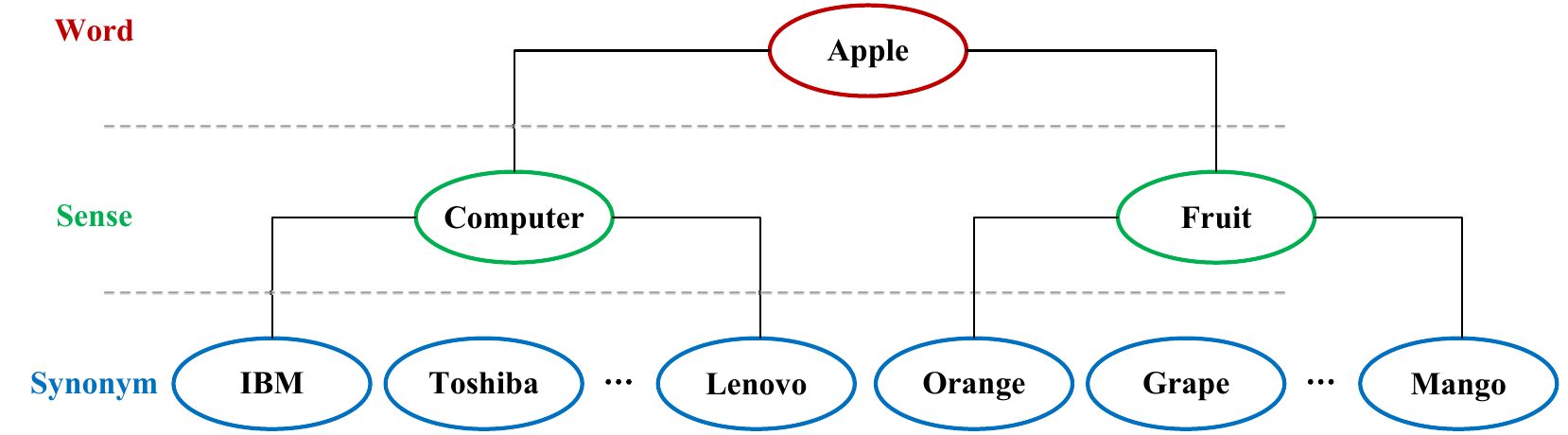}
\caption{Synonym set of the word “Apple” in OpenHowNet.} \label{synset}
\end{figure}
\noindent Furthermore, it is shown in Figure~\ref{synset} that the synonym set in OpenHowNet contains some synonym senses annotated with the same sememes annotations as the word sense. As the factors mentioned above, the synonym set of a word sense can represent its sense and have the same meaning. In this paper, we utilize the word vector of the top 10 synonyms from the pre-trained Word2Vec model and calculate the average vector of synonyms as the word vector of the word sense.
\subsection{SWSDS Model}
\begin{figure}
\centering
\includegraphics[scale=0.8]{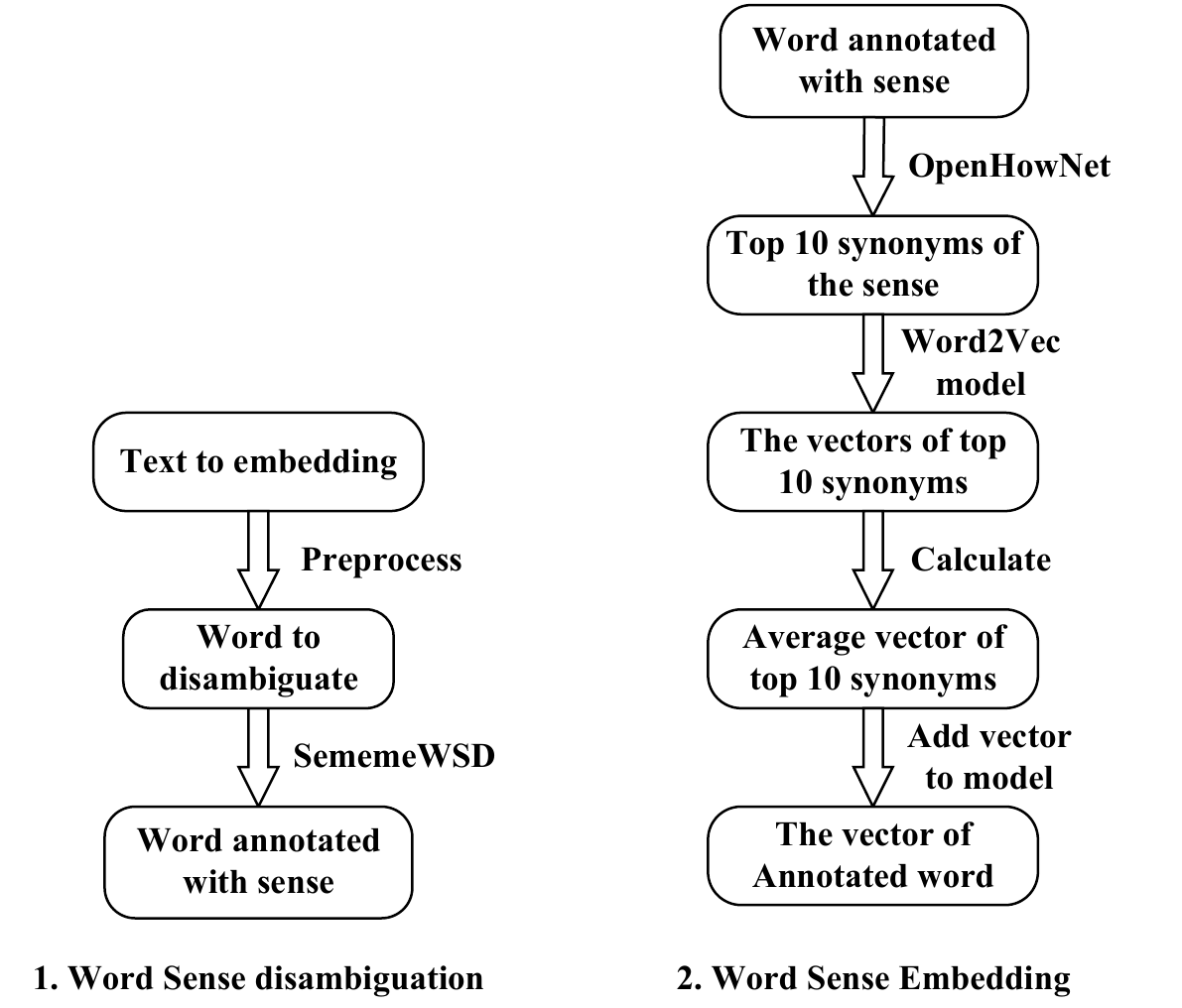}
\caption{The key process of SWSDS model.} \label{SWSDS}
\end{figure}

The architecture of our model is shown in Figure~\ref{SWSDS}. The SWSDS model can be described as a 2-stage process: word sense disambiguation and word sense embedding.

\subsubsection{Preprocess} As shown in Table~\ref{dictionary}, We build a Chinese polysemous words' dictionary to help us preprocess which has been translated into English for explanation. After preprocessing, the original text data becomes a data format that the model can process which can be shown in Table~\ref{preprocess}. 
\begin{table}
\caption{the polysemous words' dictionary.}\label{dictionary}
\centering
\begin{tabular}{|c|c|}
\hline
Full &	Verb  \\
\hline
Insist & Verb \\
\hline
Art & noun \\
\hline
... & ... \\
\hline
Question & Noun\\
\hline
\end{tabular}
\end{table}

\begin{table}
\caption{Require data format of "Apple is rich in nutrients" .}\label{preprocess}
\centering
\begin{tabular}{|c|c|}
\hline
context &	'<target>', 'is', 'rich', 'in', 'nutrients' \\
\hline
part of speech & 'n', 'v', 'a', 'p', 'n' \\
\hline
target word & Apple \\
\hline
target position & 0 \\
\hline
target word pos & n\\
\hline
\end{tabular}
\end{table}
\subsubsection{Word sense disambiguation} 
\begin{figure}
\centering
\includegraphics[scale=0.8]{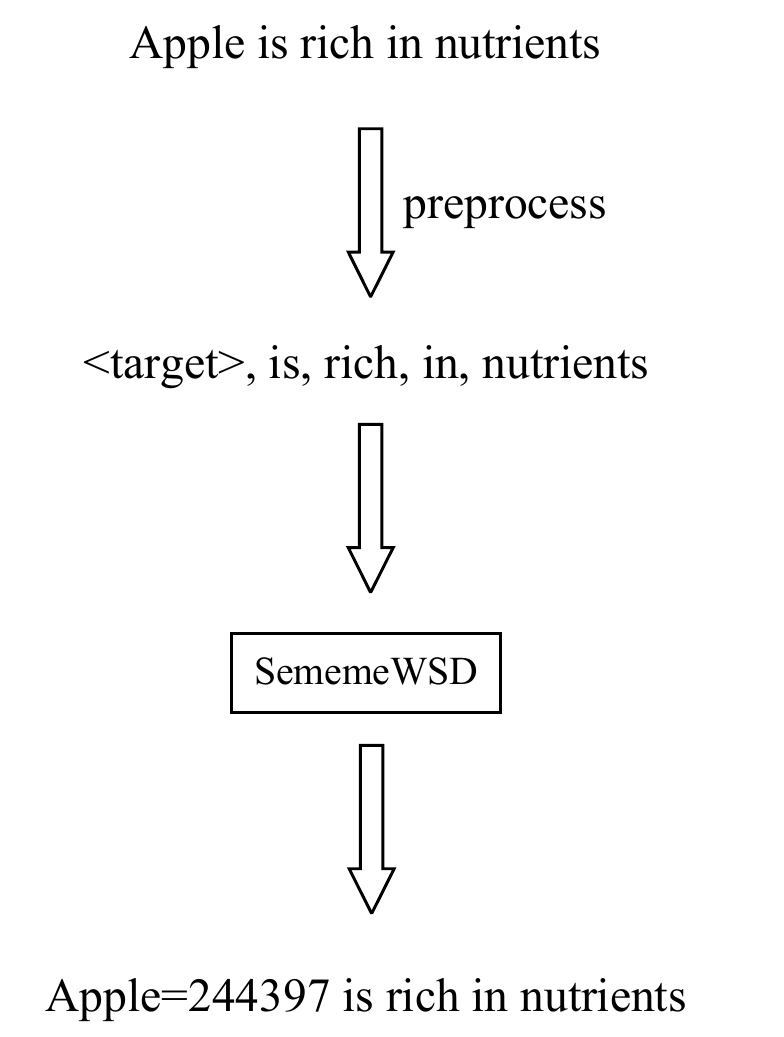}
\caption{The example of word sense disambiguation.} \label{sememewsd}
\end{figure}

Given large amounts of text data, we first preprocess text to data form that SememeWSD requires. Then, we use SememeWSD to do word sense disambiguation, and annotate polysemous words in the context with sense id. Finally, we get a sentence with annotations. For example, the polysemous word $w$“Apple” in the context $C$ “Apple is rich in nutrients” is annotated with id “244397” which means “Fruit”. That is to say, $C$ “Apple is rich in nutrients” becomes $C'$ “Apple=244397 is rich in nutrients”. The word sense disambiguation example in our model can be inllustrated in Figure~\ref{sememewsd}.
\subsubsection{Word sense embedding} Given annotated words in the context and pretrained word2vec model, we propose a simple algorithm to obtain the word vector of each annotated word: 
First, we acquire top 10 synonyms set S from OpenHowNet that have the same sememes annotation as the annotated word $w'$. Then, we obtain vectors ($V_i, i = 1...10$) of these 10 synonyms with the pre-trained word2vec model. Last, we calculate the average vector $V_{average}$ of these 10 vectors as the sense vector $V_{w'}$ of the word annotated with sense $w'$ and add it to the pretrained word2vec model. Formally, the sense vector $V_{w'}$ is computed as follows:

$$
 V_{w'} = V_{average} = \frac{1}{10}\sum_{i=1}^n V_i 
$$

\begin{figure}
\centering
\includegraphics[scale=0.8]{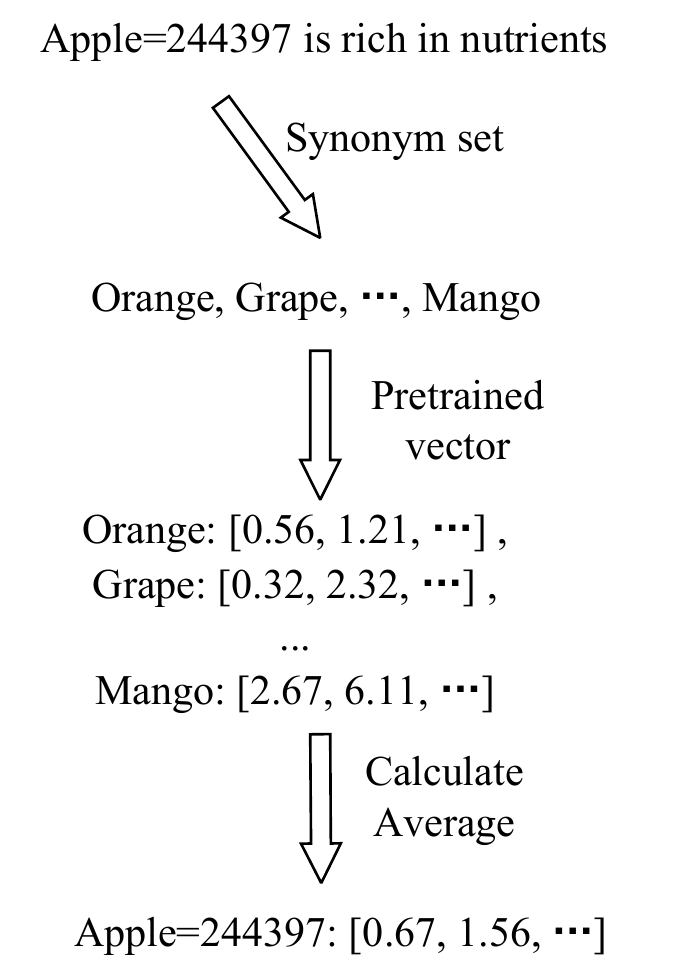}
\caption{The example of word sense embedding.} \label{wse}
\end{figure}

\noindent The calculation process is suggested in Figure~\ref{wse}. After calculation, we obtain the vector of the word sense "Apple=244397" and add it to pretrained vectors. Finally, we finish the word sense embedding and have vectors model including "Apple=244397".

\section{Experiment}
In this section, we first experiment the SememeWSD model on the OpenHowNet-based WSD dataset, showing that there are more effective BERT model for SememeWSD. Then we evaluate our SWSDS model on LCQMC(A Large scale Chinese Question Matching Corpus)~\cite{ref_article37} dataset with semantic similarity calculation.

\subsection{ SememeWSD Experiment}
We use the HowNet-based WSD dataset to evaluate different BERT models~\cite{ref_article38,ref_article39,ref_article40,ref_article41}. It is proposed by Hou et al. in their paper which is built on the Chinese Word Sense Annotated Corpus used in SemEval-2007 task 5~\cite{ref_article42}, including 2, 969 items for 36 target polysemous words (17 nouns and 19 verbs). 

Table~\ref{tab1} shows that Chinese-BERT-wwm model outperforms all the others in F1-scores for nouns words and Macro-F1 score for verbs words. Meanwhile, Chinese-
RoBERTa-wwm-ext model achieves the best performance on Micro-F1 for verbs.

\begin{table}
\caption{WSD F1 scores of Chinese BERT model.}\label{tab1}
\centering
\begin{tabular}{|c|c|c|c|c|}
\hline
\multicolumn{1}{|c|}{\multirow{2}*{Model}} &\multicolumn{2}{|c|}{Nouns} & \multicolumn{2}{|c|}{Verbs}\\
\cline{2-5}
\multicolumn{1}{|c|}{} & Micro-F1 & Macro-F1 & Micro-F1 & Macro-F1\\
\hline
Random &   38.5 & 36.1 & 23.2 & 22.8\\
\hline
Dense	&52.3	&39.0	&35.1	&33.0\\
\hline
BERT-base&	53.7&	41.7&	52.4&	48.0\\
\hline
TinyBERT&	36.0&	24.7&	20.4&	16.3\\
\hline
AlBERT-tiny	&47.9	&36.1	&38.9	&33.0\\
\hline
AlBERT-base&	51.9&	39.3&	42.3&	37.4\\
\hline
DistilBERT	&52.0&	39.1&	41.6&	35.2\\
\hline
Chinese-BERT-wwm&	54.4&	42.3&	53.6&	49.2\\
\hline
Chinese-BERT-wwm-ext&	54.0&	42.0&	53.8&	48.5\\
\hline
Chinese-RoBERTa-wwm-ext	&53.2&	40.8&	54.1&	49.2\\
\hline
\end{tabular}
\end{table}

\noindent Table~\ref{tab2} illustrates that the Chinese-BERT-wwm model achieves 53.9\% accuracy while the AlBERT-tiny model is 2.22 times faster than BERT-base model. Therefore, we use Chinese-BERT-wwm model instead of BERT-base model in experiment.
\begin{table}
\caption{WSD results and speed of Chinese BERT model.}\label{tab2}
\centering
\begin{tabular}{|c|c|c|c|c|}
\hline
\multicolumn{1}{|c|}{\multirow{2}*{Model}} &\multicolumn{3}{|c|}{All} & \multicolumn{1}{|c|}{\multirow{2}*{Speed}}\\
\cline{2-4}
\multicolumn{1}{|c|}{} & Micro-F1 & Macro-F1 & Accuracy &\\
\hline
Random	&29.1	&29.1	&29.1\%&	\\
\hline
Dense	&41.7&	35.8&	41.7\%&	\\
\hline
BERT-base&	52.9&	45.0&	53.0\%&	1x\\
\hline
TinyBERT&	26.4&	20.3&	26.4\%&	\\
\hline
AlBERT-tiny	&42.4&	34.5&	42.4\%	&2.22x\\
\hline
AlBERT-base&	46.0&	38.3&	46\%&	1.06x\\
\hline
DistilBERT&	45.6&	37.1&	45.6\%&	\\
\hline
Chinese-BERT-wwm&	53.8&	46.0&	53.9\%&	0.91x\\
\hline
Chinese-BERT-wwm-ext&	53.9&	45.5&	53.86\%	&1.01x\\
\hline
Chinese-RoBERTa-wwm-ext	&53.8&	45.2&	53.8\%&\\
\hline
\end{tabular}
\end{table}

\subsection{Semantic Similarity Calculation Evaluation}
In this section, we calculate semantic similarity with two baseline methods and SWSDS method.
\subsubsection{Experimental setting} We choose LCQMC dataset to evaluate our model. We use wmdistance method in gensim~\cite{ref_article43} library to calculate semantic similarity. The wmdistance method is implementation of the Word Mover's Distance (WMD) algorithm~\cite{ref_article44} which is a novel distance function between text documents. The WMD distance measures the semantic dissimilarity between two text documents.
\subsubsection{Baseline method} We compare our model with a high-quality word2vec model pre-trained by Tencent AILab ~\cite{ref_article45}. It provides 100-dimension vectors for over 12 million Chinese words and phrases, which are pre-trained on large-scale high-quality data.

Table~\ref{tab3} shows the results of our model and the baseline method. From the table, we can find that our model outperforms the baseline method (about 4 points higher than baseline method).
\begin{table}
\caption{Semantic similarity calculation results.}\label{tab3}
\centering
\begin{tabular}{|c|c|}
\hline
Method&	Accuracy\\
\hline
Baseline&	67.9\%\\
\hline
SWSDS&	71.9\%\\
\hline
\end{tabular}
\end{table}

\section{Conclusion}
In this paper, we present a word sense embedding method based on word sense disambiguation and synonym set. In addition, we build a polysemous words dictionary to identify polysemous words in sentences. Experimental results show that our model improves the performance of semantic similarity compared to the baseline method.

%
%

%
%
%
%

\end{document}